# Advanced Unstructured Data Processing for ESG Reports: A Methodology for Structured Transformation and Enhanced Analysis

**Jiahui Peng[1], Jing Gao[1], Xin Tong [1,*], Jing Guo[2], Hang Yang[2], Jianchuan Qi[2], Ruiqiao Li[2], Nan Li [2,†], Ming Xu[2]**

[1] Peking University, [2] Tsinghua University

[*] tongxin@urban.pku.edu.cn, [†] li-nan@tsinghua.edu.cn

**Abstract:** In the evolving field of corporate sustainability, analyzing unstructured Environmental, Social, and Governance (ESG) reports is a complex challenge due to their varied formats and intricate content. This study introduces an innovative methodology utilizing the "Unstructured Core Library", specifically tailored to address these challenges by transforming ESG reports into structured, analyzable formats. Our approach significantly advances the existing research by offering high-precision text cleaning, adept identification and extraction of text from images, and standardization of tables within these reports. Emphasizing its capability to handle diverse data types, including text, images, and tables, the method adeptly manages the nuances of differing page layouts and report styles across industries. This research marks a substantial contribution to the fields of industrial ecology and corporate sustainability assessment, paving the way for the application of advanced NLP technologies and large language models in the analysis of corporate governance and sustainability. Our code is available at https://github.com/linancn/TianGong-AI-Unstructure.git.

## 1 Introduction

In the field of data management, the dichotomy between "Structured Data" and "Unstructured Data" is critical. Structured data, typified by its presence in databases and spreadsheets, is readily processed via ETL (Extract, Transform, Load) procedures. Unstructured data, however, which includes documents, PDF files, emails, and social media posts, presents complexities due to its non-standardized formats, making context comprehension and key information extraction challenging. By 2025, it's estimated that unstructured data will constitute about 80% of the 179.6ZB of global data generation(IDC, 2023). This data is not only massive in volume but also varied in origin, posing significant challenges in extracting valuable insights due to its rapid generation and complex structure.

Particularly within this domain, ESG (Environmental, Social, and Governance) reports, predominantly in PDF format, are a notable challenge. These reports, with their blend of text, tables, and charts, offer insights into an enterprise's sustainability practices but are difficult to process due to their unstructured nature. In industrial ecology and ESG report analysis, these unstructured data are invaluable for understanding a company's environmental, social, and governance impacts. Hence, effectively processing these reports for in-depth research is imperative.

Our research explores a methodology for processing unstructured ESG report texts, making them more amenable to analysis by advanced Natural Language Processing (NLP) technologies like GPT. From intricate table data extraction to intelligent text chunking, our goal is to refine large language model responses, thereby gaining deeper insights into corporate performance in environmental, social, and governance realms. We adopt an innovative document processing strategy for Retrieval-Augmented Generation (RAG) applications(Lewis et al., 2020), using a content-aware chunking method called Unstructured. This approach allows for coherent and logical paragraph segmentation, enhancing the precision of vector databases for similarity search and large language model prompts. Our novel text processing method, based on the unstructured open-source toolkit, includes PDF content partitioning, multimodal elements integration, content segmentation, and textual content extraction. This paper details these methods and their application in processing ESG reports, aiming to improve processing results and provide tools for advanced analysis and decision support in the field of industrial ecology and sustainable business, thus offering new perspectives and tools for managing complex unstructured data in corporate settings.

## 2 Background and Related Work

Structured data, characterized by its organized and definable structure, is inherently more suitable for direct use in computer applications, as it adheres to a specific format that computer programs can easily process(Sharma and Bala, 2014). In contrast, unstructured data usually consists of information that either lacks a specific data model or has a model not readily accessible for computer programs, making it more challenging to process and analyze(Dhuria et al., 2016).

Natural Language Processing (NLP), a field at the intersection of Artificial Intelligence (AI) and linguistics, has significantly evolved since the 1960s(Lewis et al., 2020). It primarily focuses on generating and understanding natural languages, aiming to enhance human-computer interaction(Gharehchopogh and Khalifelu, 2011). In the realm of ESG report, a predominant portion of the data is unstructured. This scenario poses a significant challenge and concurrently, an area of intense research interest: the effective application of NLP techniques for the analysis and interpretation of unstructured data in ESG reports.

In recent advancements, the integration of sophisticated NLP models like BERT and GPT has catalyzed a surge in the application of diverse NLP methodologies for analyzing sustainability reports(Qiu and Jin, 2024). Amini et al. (2018) pioneered the use of Leximancer, an advanced content analysis tool, to delineate and quantify the conceptual and thematic frameworks within sustainability reports. Kang and Kim (2022) introduce an innovative methodology that addresses the limitations identified in prior works. By employing sentence similarity techniques coupled with sentiment analysis, their approach provides a nuanced understanding of thematic practices and trends in sustainability reporting, elucidating the disparity in the prevalence of positive versus negative disclosures among different corporations.

Presently, academic research in applying NLP to ESG report analysis predominantly emphasizes the development and enhancement of analytical models and methodologies. The inherent composition of ESG reports, predominantly featuring unstructured data, necessitates advanced preprocessing capabilities to mitigate potential detriments to the accuracy of subsequent analyses. This emphasis, however, often results in the underrepresentation of the inherent complexities associated with the unstructured nature of these reports. Kang and Kim (2022)selectively employed the "sent_tokenize" function from the Natural Language Toolkit (NLTK) for the dual purpose of segmenting sentences and eliminating characters not recognized by standard keyboards, thereby streamlining the textual data for subsequent analyses. However, this method can lead to instances where text is improperly segmented, resulting in a loss of accuracy in the conveyed information. Smeuninx et al. (2020) adopted ABBYY FineReader, an advanced OCR tool, to transmute report PDFs into analyzable, structured plaintext. However, this intricate process entailed manual identification and exclusion of numerical and heterogeneously composed tables, thus filtering and forwarding only relevant textual narratives to the NLP system. Reznic and Omrani (began to pay attention to the unstructured data in ESG reports, but only did basic text cleaning in the pre-processing stage.

In summation, existing methods display a marked lack of efficiency in maintaining textual sequence and coherence, coupled with a notable absence of automated, advanced techniques for extracting data from tables. Within this framework, our research has introduced a groundbreaking methodological advancement by employing the Unstructured Core Library. This innovative approach enables the effective and accurate segmentation of unstructured data within ESG reports, subsequently reorganizing it into an organized structure. This advancement effectively bridges the gaps in current methodologies, providing a refined solution that markedly improves the accuracy and efficiency of data processing in ESG reports.

# 3 Core Functions

The framework diagram in Fig.1 shows a process for organizing information from an ESG report pdf into a structured format. Initially, the pdf is split into text, images, tables, headers and footers. The headers and footers are filtered, the text is cleaned up, images are textualized, and tables are converted to HTML. These elements are then grouped together into sections, each with a title and body, creating an organized list of the report's contents in a clear, structured form.

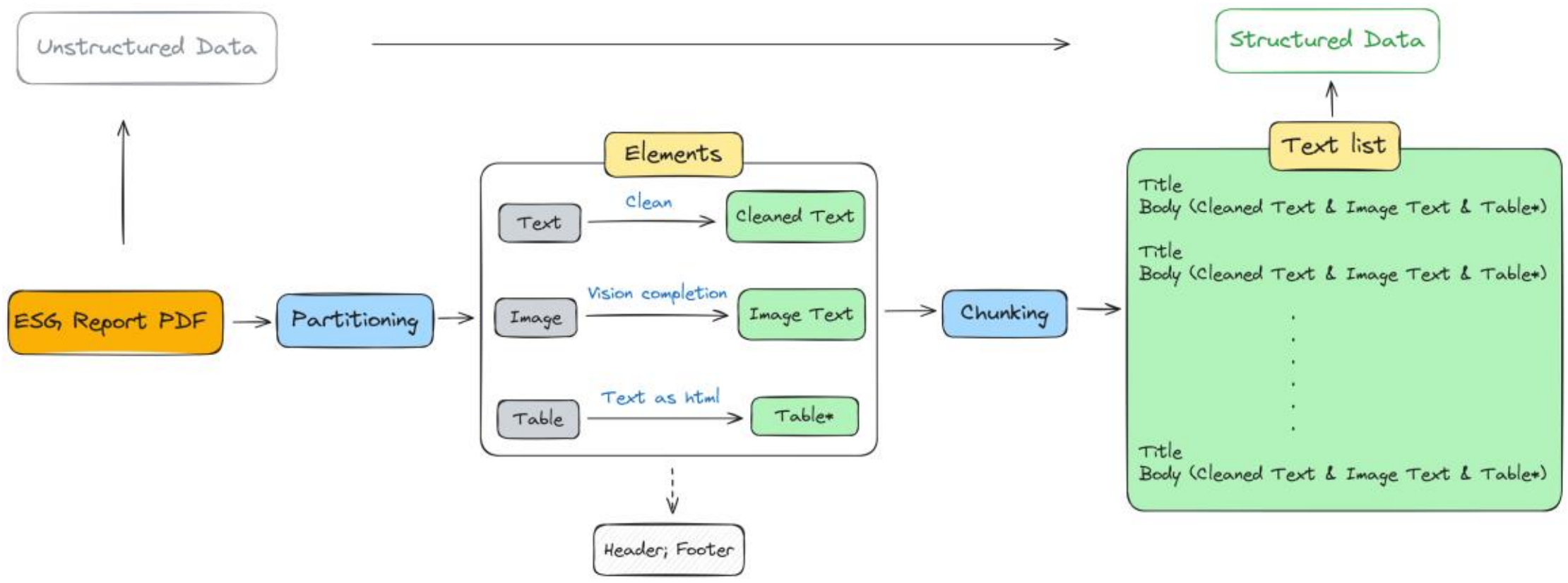

Fig. 1 Unstructured processing framework

### 3.1 Partitioning

During the Partitioning process, the ESG report pdf is divided into multiple elements including texts, images, tables, headers and footers with the "hi_res" strategy(Unstructured, 2023), utilizing detectron2, a deep learning-based object detection system. This strategy is particularly advantageous for its ability to leverage the layout information to gain additional insights about the document elements, not just the text but also the formatting and structure like headings, paragraphs, and tables. A notable limitation of the "hi_res" strategy is its difficulty in correctly ordering elements in multi-column documents, potentially affecting the proper sequencing of content. Overall, the "hi_res" strategy offers a comprehensive approach to pdf processing, particularly effective for complex layouts but dependent on the availability of specific technologies like detectron2.

### 3.2 Multimodal Elements Integration and Standardization

The post-partitioning elements require subsequent refinement, purification, and structuring. This process involves the elimination of ancillary components such as "Headers" and "Footers", while retaining standardized "Text", "Image", and "Table" data.

Initially, the procedure selectively excludes "Headers" and "Footers" from the elements, preserving solely three elements: "Text", "Image", and "Table". For the "Text" elements, meticulous cleansing is imperative prior to their integration into the NLP model, to mitigate potential detrimental impacts on model efficiency caused by superfluous content. Addressing this challenge, the "Unstructured Documentation" cleaning functions are employed(Unstructured, 2023). These utilities proficiently consolidate paragraphs demarcated by newlines, excise bullets and dashes at text commencement, and eradicate excessive whitespace, thereby enhancing the clarity and integrity of the textual data.

Concerning "Image" elements, the "vision_completion" technique is employed. This method involves the activation of the "gpt-4-vision-preview" API, whereby tailored queries are formulated to procure descriptive textual interpretations from GPT about the image's contents. Subsequently, this generated textual output is re-integrated into the dataset, precisely at the original position of the corresponding image.

For "Table" elements, the raw text of the table will be stored in the text attribute for the element, and the HTML representation of the table will be available in the element's metadata under "element.metadata.text_as_html", so it is output in that form to preserve the formatting of the table(Unstructured, 2023).

## 3.3 Chunking

The "Unstructured Core Library" chunking functions are crucial for document processing, particularly in applications like RAG(Unstructured, 2023). A key function, "chunk_by_title", segments documents into subsections by detecting titles, each title commencing a new section. Notably, specific elements such as tables and non-text components (e.g., page breaks, images) inherently constitute separate sections.

Activating the "multipage_sections" parameter facilitates the creation of sections spanning multiple pages. This feature is essential for preserving thematic or contextual continuity across page breaks, thus ensuring the segmented document mirrors the structure of the original text.

The "combine_text_under_n_chars" parameter, set to 0, overrides the default behavior of merging consecutive text sections below a certain character limit (commonly 500 characters). This setting guarantees that all text sections are recognized as individual segments, maintaining the document's structural granularity.

The "new_after_n_chars" parameter remains unspecified (set to None), indicating reliance on the function's inherent setting for initiating new sections. Meanwhile, the "max_characters" parameter, fixed at 4096, implies an adaptation to include larger sections within each chunk, catering to the document's specific needs.

## 3.4 Structured data integration

This process entails the systematic handling of segmented document chunks to classify and structure text and table elements into an organized data format. The procedure iterates over each chunk, extracting text from instances of "Composite Element" and accumulating it in a list. For elements categorized as "Table", their HTML representations are either integrated into this list or merged with the preceding text to maintain a uniform compilation of related content. This operation yields a series of text strings, each potentially encompassing a title and its corresponding content body. Subsequently, these text strings are dissected into title-body duos using a specific delimiter. Each duo is cataloged in a dictionary, with distinct keys assigned for the title and body. These dictionaries are then assembled into a list, representing the methodically organized content. The final phase involves converting this structured content into a pandas Data Frame, subsequently exported to an Excel file. This conversion is pivotal for enabling further data analysis and manipulation, effectively transforming the original unstructured document segments into an efficiently organized, tabular representation.

# 4 Experiments and Discussion

Three Fortune 500 companies, each representing a distinct industry classification - Walmart, Apple, and Toyota - were selected for analysis（Table.1）. Their latest ESG reports underwent a comprehensive unstructured split processing. This involved a detailed examination of the reports' text, images, and tables. The primary aim of this empirical study was to assess the effectiveness of unstructured processing techniques in analyzing ESG reports from diverse industry sectors. The findings are detailed below:

Table. 1 Basic company information

| Company | Country | Industry |
|---|---|---|
| Walmart | United States | Retail |
| Apple | United States | Technology |
| Toyota | Japan | Automotive |

## 4.1 Text processing results

In demonstrating the efficacy of our proposed method, we selected three distinct text blocks from diverse industries and formatting styles: Walmart FY2023 HIGHLIGHTS, Apple's Ambitious goals, and Toyota ESD Project(Table.2). The processed results are visibly discernible through "Title" and "Cleaned Text". This methodology adeptly identifies "Title" across varied ESG report formats from these distinct sectors and executes thorough text cleaning and organization. Notably, in the Apple report, key themes such as "Climate Change", "Resources", and "Smarter Chemistry" were precisely identified, exemplifying the method's robustness in handling heterogeneous data structures and themes.

## 4.2 Image processing results

The image processing results (Table.3), as illustrated, indicate that for the majority of images in ESG reports, crucial information is predominantly stored in textual format within the images, enabling direct text extraction. Consequently, the current functionality of our method is tailored to process images that do not contain embedded text.

## 4.3 Table processing results

The processing of tables represents the most challenging aspect of our work, a facet previously overlooked in prior research (Table.4). The post-processing results of tables demonstrate that they have been organized into concise and clear formats using the "text_as_html" method. Specifically, for Walmart's Original Table, a less conspicuous table element was successfully identified and processed as a table. Apple's Original Table, considerably more complex with 21 rows and 7 columns, underwent meticulous segmentation and organization. However, due to the physical separation of texts like "Water", "Waste", and "Product packaging footprint" from the core table area, these elements were not captured. And the unstructured performance in the upper part of the table is not satisfactory, but it's possible that large language models might still be able to understand it. Toyota's Original Table, being more standardized, yielded a distinctly clear and well-processed result.

Table. 2 Walmart, Apple, Toyota text processing results

<table>
<tr><th>Company</th><th>Original Text</th><th>Title</th><th>Cleaned Text</th></tr>
<tr><td>Walmart</td><td>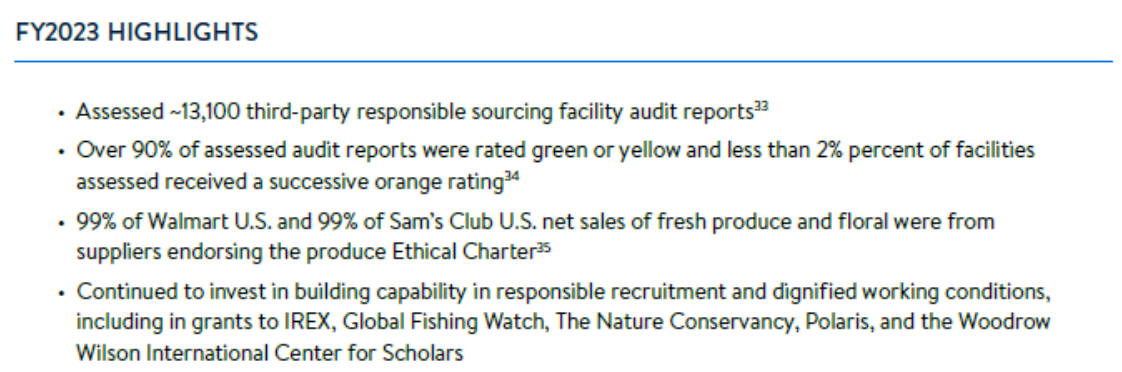<br>FY2023 HIGHLIGHTS<br>• Assessed ~13,100 third-party responsible sourcing facility audit reports[33]<br>• Over 90% of assessed audit reports were rated green or yellow and less than 2% percent of facilities assessed received a successive orange rating[34]<br>• 99% of Walmart U.S. and 99% of Sam's Club U.S. net sales of fresh produce and floral were from suppliers endorsing the produce Ethical Charter[35]<br>• Continued to invest in building capability in responsible recruitment and dignified working conditions, including in grants to IREX, Global Fishing Watch, The Nature Conservancy, Polaris, and the Woodrow Wilson International Center for Scholars</td><td>FY2023 HIGHLIGHTS</td><td>Assessed ~13,100 third party responsible sourcing facility audit reports33<br><br>Over 90% of assessed audit reports were rated green or yellow and less than 2% percent of facilities assessed received a successive orange rating34<br><br>99% of Walmart U.S. and 99% of Sam's Club U.S. net sales of fresh produce and floral were from suppliers endorsing the produce Ethical Charter35<br><br>Continued to invest in building capability in responsible recruitment and dignified working conditions, including in grants to IREX, Global Fishing Watch, The Nature Conservancy, Polaris, and the Woodrow Wilson International Center for Scholars</td></tr>
<tr><td rowspan="3">Apple</td><td rowspan="3">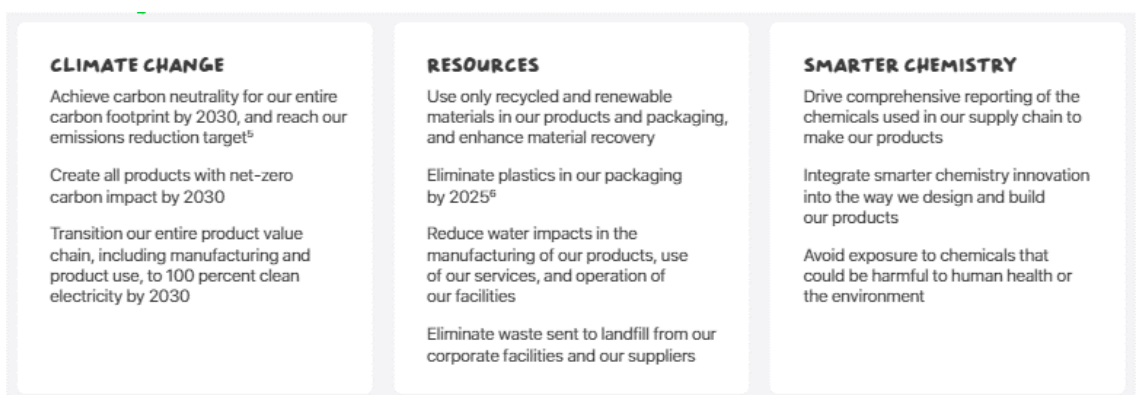<br>CLIMATE CHANGE<br>Achieve carbon neutrality for our entire carbon footprint by 2030, and reach our emissions reduction target[5]<br>Create all products with net-zero carbon impact by 2030<br>Transition our entire product value chain, including manufacturing and product use, to 100 percent clean electricity by 2030<br>RESOURCES<br>Use only recycled and renewable materials in our products and packaging, and enhance material recovery<br>Eliminate plastics in our packaging by 2025[6]<br>Reduce water impacts in the manufacturing of our products, use of our services, and operation of our facilities<br>Eliminate waste sent to landfill from our corporate facilities and our suppliers<br>SMARTER CHEMISTRY<br>Drive comprehensive reporting of the chemicals used in our supply chain to make our products<br>Integrate smarter chemistry innovation into the way we design and build our products<br>Avoid exposure to chemicals that could be harmful to human health or the environment</td><td>Climate Change</td><td>Achieve carbon neutrality for our entire carbon footprint by 2030, and reach our emissions reduction target5<br><br>Create all products with net zero carbon impact by 2030<br><br>Transition our entire product value chain, including manufacturing and product use, to 100 percent clean electricity by 2030</td></tr>
<tr><td>Resources</td><td>Use only recycled and renewable materials in our products and packaging, and enhance material recovery<br><br>Eliminate plastics in our packaging by 20256<br><br>Reduce water impacts in the manufacturing of our products, use of our services, and operation of our facilities<br><br>Eliminate waste sent to landfill from our corporate facilities and our suppliers</td></tr>
<tr><td>Smarter Chemistry</td><td>Drive comprehensive reporting of the chemicals used in our supply chain to make our products</td></tr>
</table>

| | | | |
|---|---|---|---|
| | | | Integrate smarter chemistry innovation into the way we design and build our products<br><br>Avoid exposure to chemicals that could be harmful to human health or the environment |
| Toyota | **— Toyota ESD Project —**<br>**Environmental education for the next generation ⇒ "Connecting to the Future" activities**<br><br>2025 Target<br>● Implement globally unified initiatives to foster environmentally conscious persons responsible for the future.<br>● Offer environmental education opportunities by utilizing biotopes and others in collaboration with the Plant in Harmony with Nature.<br>● Foster environmentally conscious persons at both in-house and outside sites, including plants and the Forest of Toyota, by utilizing educational tools in harmony with nature for the next generation. | - Toyota ESD Project - Environmental education for the next generation ⇒ "Connecting to the Future" activities | 2025 Target Implement globally unified initiatives to foster environmentally conscious persons responsible for the future.<br><br>Offer environmental education opportunities by utilizing biotopes and others<br><br>in collaboration with the Plant in Harmony with Nature.<br><br>Foster environmentally conscious persons at both in house and outside sites, including plants and the Forest of Toyota, by utilizing educational tools in harmony with nature for the next generation. |

Table. 3 Walmart, Apple, Toyota image processing results

| Company | Original Image | Image Text |
|---|---|---|
| Walmart | 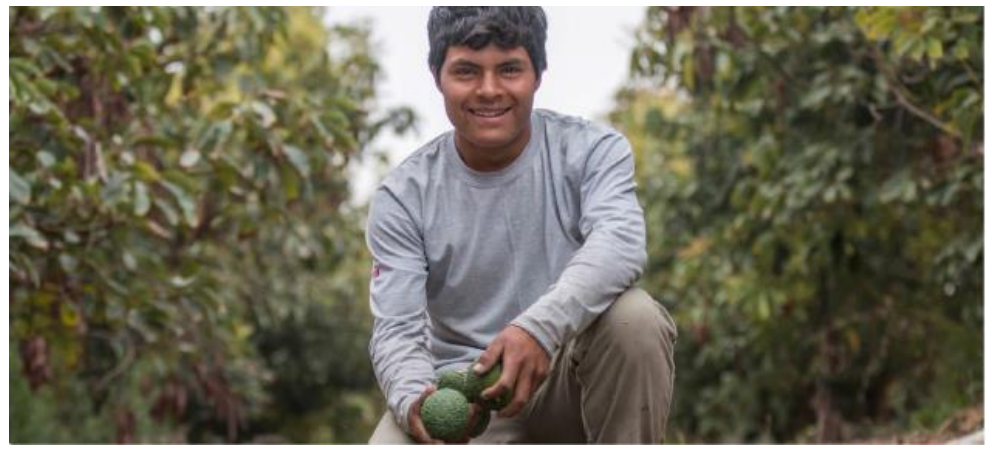 | The image depicts a person outdoors, smiling at the camera and holding an avocado. They appear to be in a natural environment suggestive of an avocado farm, given the lush greenery and the appearance of avocado trees in the background. The individual is dressed in casual attire, which is practical for agricultural work, and appears to be either harvesting or inspecting the fruit, signifying involvement in farming or agricultural practices. This scene conveys a sense of agricultural life, potentially emphasizing the human aspect of farming and the cultivation of produce. |
| Apple | 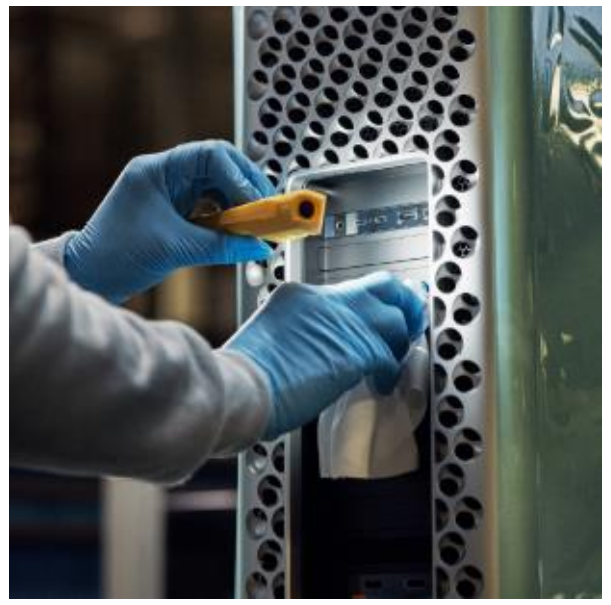 | This image shows a person wearing blue gloves engaging in what appears to be a cleaning or maintenance activity on a piece of electronic equipment, which seems to be a computer server or data storage unit. The individual is using a yellow tool, possibly a cutter or prying device, and a white cloth, which could indicate that they are either cleaning sensitive components or preparing to perform hardware maintenance tasks such as opening the device or ensuring a dust-free environment. The compartmentalized design of the metal surface with circular perforations suggests that this is part of a system designed for efficient airflow, which is essential for cooling in high-performance computing equipment. |
| Toyota | 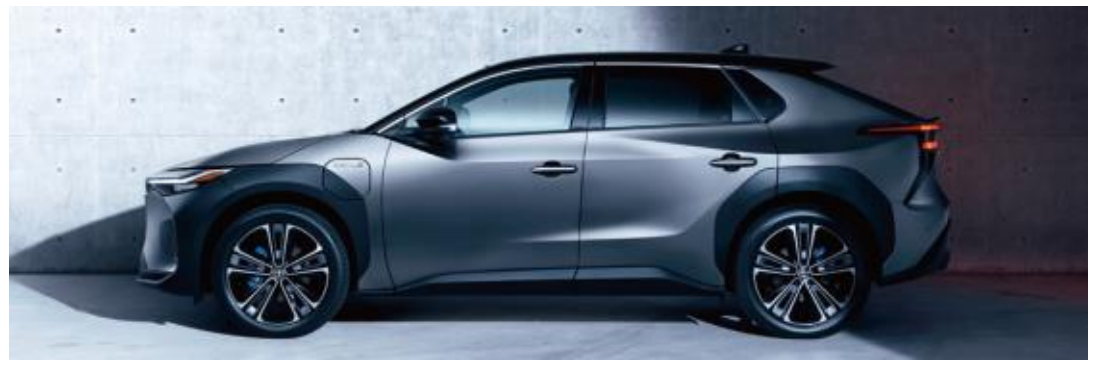 | The image is of a modern SUV. It exhibits a coupe-like roofline and features sporty alloy wheels, which are common design choices aimed at combining utility with a stylish aesthetic. The paint appears to be a matte or satin finish, which is a trend that has become more popular in recent years. The bodywork also shows pronounced creases and the door handles are flush with the body, indicating a focus on aerodynamics. Large wheels and low-profile tires are indicative of an emphasis on performance in addition to everyday utility. The lack of visible exhaust tips could suggest that this vehicle is an electric or hybrid model, aligning with the increasing focus on sustainability in the automotive industry. |

Table. 4 Walmart, Apple, Toyota table processing results

| Company | Original Table | Table presented in html |
|---|---|---|
| Walmart |  |  |
| Apple |  |  |
| Toyota |  |  |

**Walmart — Original Table**

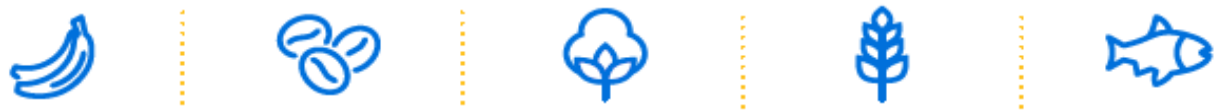

| Floral & Produce | Specialty Crops | Textile, Pulp, Paper & Timber | Ingredients | Proteins |
|---|---|---|---|---|
| Fresh flowers, apples, bananas, berries, grapes, leafy greens, pineapples, stone fruit, and tomatoes | Cocoa, coffee, and tea | Cotton textiles, pulp, paper, and timber | U.S. corn, U.S. wheat, U.S. soy, South American soy, and palm oil | Seafood (wild caught and farm raised), meat (South American beef, U.S. beef, pork, and poultry), and dairy |

**Walmart — Table presented in html**

| Floral & Produce | Specialty Crops | Textile, Pulp, | Ingredients | Proteins |
|---|---|---|---|---|
| Fresh flowers, apples, bananas, berries, grapes, leafy greens, pineapples, stone fruit, and tomatoes | Cocoa, coffee, and tea | Paper & Timber Cotton textiles, pulp, and paper, timber | U.S. corn, U.S. wheat, U.S. soy, South American soy, and palm oil | Seafood (wild caught and farm raised), meat (South Americar beef, U.S. beef, pork, and poultry and dairy |

**Apple — Original Table**

| | | Unit | Fiscal year 2022 | 2021 | 2020 | 2019 | 2018 |
|---|---|---|---|---|---|---|---|
| Water | Corporate facilities | | | | | | |
| | Total | million gallons | 1,527 | 1,407 | 1,287 | 1,291 | 1,258 |
| | Freshwater[1] | million gallons | 1,380 | 1,259 | 1,168 | 1,178 | 1,190 |
| | Recycled water[2] | million gallons | 142 | 141 | 113 | 106 | 63 |
| | Other alternative sources[3] | million gallons | 5 | 7 | 5 | 7 | 4 |
| | Supply chain | | | | | | |
| | Freshwater saved | million gallons | 13,000 | 12,300 | 10,800 | 9,300 | 7,600 |
| Waste | Corporate facilities[4] | | | | | | |
| | Landfill diversion rate | % | 71 | 68 | 70 | 66 | 67 |
| | Landfilled (municipal solid waste) | pounds | 33,260,990 | 33,202,200 | 25,826,550 | 38,317,120 | 32,372,890 |
| | Recycled | pounds | 78,618,250 | 73,489,220 | 63,812,300 | 72,338,130 | 66,380,630 |
| | Composted | pounds | 8,726,170 | 4,844,960 | 6,302,410 | 10,882,120 | 10,397,430 |
| | Hazardous waste | pounds | 2,780,610 | 3,525,840 | 4,053,770 | 6,096,600 | 6,277,800 |
| | Waste to energy | pounds | 1,197,570 | 657,890 | 786,250 | 1,129,080 | 1,105,140 |
| | Supply chain | | | | | | |
| | Waste diverted from landfill | metric tons | 523,000 | 491,000 | 400,000 | 322,000 | 375,000 |
| Product packaging footprint | Packaging | | | | | | |
| | Total | metric tons | 276,100 | 257,000 | 226,000 | 189,000 | 187,000 |
| | Recycled fiber | % of total | 66 | 63 | 60 | 59 | 58 |
| | Responsibly sourced virgin fiber[5] | % of total | 30 | 33 | 35 | 33 | 32 |
| | Plastic | % of total | 4 | 4 | 6 | 8 | 10 |

**Apple — Table presented in html**

| | | | | | | | | | |
|---|---|---|---|---|---|---|---|---|---|
| | | Unit | 2022 | 2021 | | 2019 | | | |
| | | Corporate facilities | | | | | | | |
| | | Total | million gallons | 1,527 | | 1,407 | | | 1,291 |
| | | Freshwater! | million gallons | 1,380 | 2020 1,287 1168 113 5 | 1,259 | | | 1178 |
| er | | Recycled water? | million gallons | 142 | | 141 | 2018 1,258 1,190 63 4 7,600 | | 106 |
| | | Other alternative sources? | million gallons | 5 | | 7 | | | 7 |
| | | Supply chain | | | | | | | |
| | | Freshwater saved | million gallons | 13,000 | 12,300 | 10,800 | | 9,300 | |
| | Corporate facilities* | | | | | | | | |
| | | Landfill diversion rate | % | 71 | 68 | 70 | 66 | 67 | |
| | | Landfilled (municipal solid waste) | pounds | 33,260,990 | 33,202,200 | 25,826,550 | 38,317,120 | 32,372,890 | |
| te | | Recycled | pounds | 78,618,250 | 73,489,220 | 63,812,300 | 72,338,130 | 66,380,630 | |
| | | Composted | pounds | 8,726,170 | 4,844,960 | 6,302,410 | 10,882,120 | 10,397,430 | |
| | | Hazardous waste | pounds | 2,780,610 | 3,525,840 | 4,053,770 | 6,096,600 | 6,277,800 | |
| | | Waste to energy | pounds | 1,197,570 | 657,890 | 786,250 | 1,129,080 | 1,105,140 | |
| | | Supply chain | | | | | | | |
| | | Waste diverted from landfill | metric tons | 523,000 | 491,000 | 400,000 | 322,000 | 375,000 | |
| | Packaging | | | | | | | | |
| | | Total | metric tons | 276,100 | 257,000 | 226,000 | 189,000 | 187,000 | |
| Juct packaging print | | Recycled fiber | % of total | 66 | 63 | 60 | 59 | 58 | |
| | | Responsibly sourced virgin fiber> | % of total | 30 | 33 | 35 | 33 | 32 | |
| | | Plastic | % of total | 4 | 4 | 6 | 8 | 10 | |

**Toyota — Original Table**

(million t-$CO_2e$)

| By type | 2019 | 2020 | 2021 |
|---|---|---|---|
| Non-energy-related $CO_2$ | 0.008 | 0.007 | **0.007** |
| $CH_4$ | 0.015 | 0.015 | **0.013** |
| $N_2O$ | 0.009 | 0.008 | **0.009** |
| PFCs | 0.009 | 0.008 | **0.041** |
| HFCs | 0 | 0 | **0** |
| $SF_6$ | 0.002 | 0.005 | **0.002** |
| Total | 0.042 | 0.043 | **0.072** |

Calculated in accordance with the Japanese Act on Promotion of Global Warming Countermeasures

**Toyota — Table presented in html**

| By type | 2019 | 2020 | 2021 |
|---|---|---|---|
| Non-energy-related COz | 0.008 | 0.007 | 0.007 |
| CH4 | 0.015 | 0.015 | 0.013 |
| N2O | 0.009 | 0.008 | 0.009 |
| PFCs | 0.009 | 0.008 | 0.041 |
| HFCs | 0 | 0 | 0 |
| SFe | 0.002 | 0.005 | 0.002 |
| Total | 0.042 | 0.043 | 0.072 |
| alculated in accordance with the Japanese Act on Promotion of Global Warmin ountermeasures | | | |

## 5 Conclusion

This study presents a groundbreaking approach in processing and integrating unstructured data from ESG reports, utilizing advanced techniques to convert these complex documents into structured, analyzable formats. Our methodology, underpinned by the Unstructured library, demonstrates a significant advancement over existing methods, particularly in handling the diverse elements of PDFs such as text, images, and tables.

We've shown through empirical analysis that our method can adeptly manage various data types found in ESG reports, ensuring that crucial information is not only preserved but also made accessible for in-depth analysis. The method excels in processing text by effectively identifying and cleaning titles and body sections, as demonstrated in reports from Walmart, Apple, and Toyota. Additionally, it addresses the challenge of image processing by focusing on images devoid of embedded text, a common feature in ESG reports.

One of the most significant achievements of our approach is its capability to handle complex table structures, a task that has been notably challenging in previous studies. Our method's ability to reorganize tables into a more structured and coherent format enhances the readability and analysis of these critical data elements.

It is imperative to acknowledge certain limitations inherent in the current research. Despite the methodology's robust performance in structuring and analyzing unstructured data from ESG reports, challenges persist due to the considerable variability in report formats and writing styles across different industries and corporations. Diverse page layouts can influence the efficacy of data segmentation, though, on the whole, the method succeeds in achieving clear and precise data division. Moreover, a notable deficiency of our approach lies in its handling of data charts embedded with text. The current methodology does not optimally process these data-rich visual elements, which remains an area for future enhancement.

Despite these limitations, the study represents a significant advancement in processing unstructured data from ESG reports, offering a substantial contribution to the fields of industrial ecology and corporate sustainability assessment. The insights and methods developed in this study pave the way for future research to refine and expand upon these initial findings, particularly in the realm of integrating complex data visualizations and further enhancing the adaptability of the methodology to diverse report structures.

## Acknowledgements

This research was supported by one of the projects of the Erasmus Initiative: Dynamics of Inclusive Prosperity, a joint project funded by the Dutch Research Council (NWO) and the National Natural Science Foundation of China (NSFC): “Towards Inclusive Circular Economy: Transnational Network for Wise-waste Cities (IWWCs)” (NSFC project number: 72061137071；NWO project number: 482.19.608), and the project of " ESG Text Analysis for Assessing Carbon Disclosure Quality Based on Large Language Models" funded by the SMP (Social Media Processing)-ZHIPU AI Large Language Model Interdisciplinary Fund.